\PassOptionsToPackage{table}{xcolor}
\documentclass[11pt,a4paper]{article}
\usepackage[hyperref]{emnlp2018}
\pdfoutput=1

\usepackage{xcolor}
\usepackage{times}
\usepackage{latexsym}
\usepackage{amsmath}
\usepackage{algorithm}
\usepackage{amsfonts}
\usepackage{lipsum}
\usepackage{multirow}
\usepackage{framed}
\usepackage{graphicx}
\usepackage{longtable}
\usepackage{caption}
\usepackage{subcaption}
\usepackage{graphics}
\usepackage{booktabs}
\usepackage[flushleft]{threeparttable}
\usepackage{tabularx}
\usepackage{tikz}
\usepackage{bm}
\usepackage[all]{nowidow}

\usepackage{tikz}
\usepackage{xcolor}
\definecolor{color0}{HTML}{0000FF}
\definecolor{color1}{HTML}{3232FF}
\definecolor{color2}{HTML}{6666FF}
\definecolor{color3}{HTML}{9898FF}
\definecolor{color4}{HTML}{CCCCFF}
\definecolor{color5}{HTML}{FFFEFE}
\definecolor{color6}{HTML}{FFCCCC}
\definecolor{color7}{HTML}{FF9898}
\definecolor{color8}{HTML}{FF6666}
\definecolor{color9}{HTML}{FF3232}
\newcommand*{\mybox}[2]{\tikz[anchor=base,baseline=0pt,rounded corners=0pt, inner sep=0.2mm] \node[fill=#1] (X) {#2};}

\definecolor{red2}{RGB}{215,25,28}
\definecolor{red1}{RGB}{243, 164, 129}
\definecolor{red0}{RGB}{253, 217, 196}
\definecolor{blue2}{RGB}{42, 113, 178}
\definecolor{blue1}{RGB}{123, 182, 214}
\definecolor{blue0}{RGB}{213, 231, 241}

\definecolor{pink2}{RGB}{213, 87, 157}
\definecolor{pink1}{RGB}{238, 173, 212}
\definecolor{pink0}{RGB}{248, 208, 231}
\definecolor{green2}{RGB}{67, 134, 31}
\definecolor{green1}{RGB}{127, 188, 65}
\definecolor{green0}{RGB}{199, 232, 159}
\usepackage{ragged2e}
\newcolumntype{Y}{>{\RaggedRight\arraybackslash}X}

\newcommand{\abr}[1]{\textsc{#1}}
\newcommand{\dknn}{\textsc{DkNN}}
\newcommand{\lstm}{\textsc{LSTM}}
\newcommand{\bilstm}{\textsc{BiLSTM}}
\newcommand{\cnn}{\textsc{CNN}}
\newcommand{\loo}{leave-one-out}
\newcommand{\mb}[1]{\boldsymbol{\mathbf{#1}}}

\urldef{\footurl}\url{https://sites.google.com/view/language-dknn/}

\newif\ifcomment\commenttrue
\ifcomment
 \newcommand{\jbgcomment}[1]{ \colorbox{red}{   \parbox{.8\linewidth}{ JBG: #1}  }}
 \newcommand{\fscomment}[1]{ \colorbox{green}{   \parbox{.8\linewidth}{ fs: #1}  }}
 \newcommand{\ewcomment}[1]{ \colorbox{blue}{   \parbox{.8\linewidth}{ Eric: #1}  }}
\else
\newcommand{\jbgcomment}[1]{}
\newcommand{\fscomment}[1]{}
\newcommand{\ewcomment}[1]{}
\fi

\graphicspath{{2018_emnlp_knn/images/}, {2018_emnlp_knn/auto_fig/}}

\aclfinalcopy

\makeatletter
\g@addto@macro{\UrlBreaks}{\do\/\do\-}
\makeatother

\makeatletter
\def\BState{\State\hskip-\ALG@thistlm}
\newcommand{\printfnsymbol}[1]{%
  \textsuperscript{\@fnsymbol{#1}}%
}
\makeatother

\title{Interpreting Neural Networks With Nearest Neighbors}

\author{Eric Wallace\thanks{$^{\star}$Equal contribution} , Shi Feng\printfnsymbol{1}, Jordan Boyd-Graber\\
University of Maryland \\
  {\tt \{ewallac2,shifeng,jbg\}@umiacs.umd.edu} \\}
\date{}

\begin{document}
\maketitle

\begin{abstract}
    Local model interpretation methods explain individual predictions by
assigning an importance value to each input feature. This value is
often determined by measuring the change in 
confidence when a feature is removed. However, the confidence of neural networks 
is not a robust measure of model uncertainty. This issue makes 
reliably judging the importance of the input features difficult. 
We address this by changing the test-time behavior of neural networks
using Deep k-Nearest Neighbors. Without harming text classification accuracy, 
this algorithm provides a more robust uncertainty
metric which we use to generate feature importance values.
The resulting interpretations better align with human perception than
baseline methods. Finally, we use our interpretation method
to analyze model predictions on
dataset annotation artifacts.
\end{abstract}

\section{Introduction}
\label{sec:introduction}

The growing use of neural networks in sensitive domains such as
medicine, finance, and security raises concerns about human
trust in these machine learning systems. A central question is
test-time \emph{interpretability}: how can humans understand the
reasoning behind model predictions?

A common way to interpret neural network predictions is to identify the
most important input features. For
instance, a saliency map that highlights important pixels in an
image~\cite{sundararajan2017axiomatic} or words in a
sentence~\cite{li2016understanding}. Given a test prediction,
the \emph{importance} of each input feature is the change in model confidence when that
feature is removed.

However, neural network confidence is not a proper measure
of model uncertainty~\cite{guo2017calibration}. This issue is emphasized
when models make highly confident predictions on inputs that
are completely void of information, for example, images of pure
noise~\cite{goodfellow2014explaining} or meaningless text
snippets~\cite{feng2018rawr}. Consequently, a model's
confidence may not properly reflect whether discriminative
input features are present. This issue makes it difficult 
to reliably judge the importance of each input feature using common
confidence-based interpretation methods~\cite{feng2018rawr}.

To address this, we apply Deep k-Nearest Neighbors
(\dknn{})~\cite{papernot2018dknn} to neural models for text
classification. Concretely, predictions are no longer
made with a softmax classifier, but using
the labels of the \emph{training examples} whose representations are
most similar to the test example (Section~\ref{sec:deepknn}).
This provides an alternative metric for
model uncertainty, \emph{conformity}, which measures how much
support a test prediction has by comparing its hidden representations
to the training data. This representation-based uncertainty
measurement can be used in combination with existing 
interpretation methods, such as \loo{}~\cite{li2016understanding}, to better identify important
input features.

We combine \dknn{} with \cnn{} and \lstm{} models on six
\abr{nlp} text classification tasks, including sentiment analysis and textual entailment, with no loss
in classification accuracy (Section~\ref{sec:experiments}). We compare interpretations generated using \dknn{}
conformity to baseline interpretation methods, finding \dknn{} interpretations rarely assign
importance to extraneous words that do not align with human perception (Section~\ref{sec:interpretation}).
Finally, we generate interpretations using \dknn{} conformity for a dataset with known artifacts (\abr{snli}),
helping to indicate whether a model has learned superficial patterns.
We open source the code for \dknn{} and our results.\footnote{https://github.com/Eric-Wallace/deep-knn}

\section{Interpretation Through Feature Attribution}
\label{sec:saliency_approaches}

Feature attribution methods explain a test prediction
by assigning an importance value to each input
feature (typically pixels or words).

In the case of text classification, we
have an input sequence of $n$ words $\mb{x}=\langle w_1, w_2, \ldots
w_n\rangle$, represented as
one-hot vectors. The word sequence is then converted to a sequence of
word embeddings~$\mb{e}=\langle\bm{v}_1, \bm{v}_2, \ldots \bm{v}_n\rangle$.
A classifier $f$
outputs a probability distribution over classes.
The class with the highest probability is selected as the prediction $y$, with its
probability serving as the model confidence. To create an interpretation, each input word
is assigned an importance value, $g(w_i \mid \mb{x}, y)$, which indicates the word's
contribution to the prediction. A saliency map (or heat map)
visually highlights the words in a sentence.

\subsection{Leave-one-out Attribution}
\label{sec:l10}
A simple way to define the importance $g$ is via \loo{}~\cite{li2016understanding}: individually
remove a word from the input and see how the confidence changes. The importance of
word $w_i$ is the decrease in confidence\footnote{equivalently the change in class score or cross
entropy loss} when word $i$ is removed:
\begin{equation}
\label{eq:importance}
    g(w_i \mid \mb{x}, y) = f(y\mid\mb{x}) - f(y\mid\mb{x}_{-i}),
\end{equation}
where $\mb{x}_{-i}$ is the input sequence with the $i$th word removed and
$f(y\mid\mb{x})$ is the model confidence for class $y$. This can be repeated for
all words in the input.
Under this definition, the sign of the importance value is
opposite the sign of the confidence change: if a word's removal 
causes a decrease in the confidence, it gets a positive
importance value. 
We refer to this interpretation method as \textit{Confidence} \loo{}
in our experiments.

\subsection{Gradient-Based Feature Attribution}
\label{sec:gradient}
In the case of neural networks, the model $f(\mb{x})$ as a function of word
$w_i$ is a highly non-linear, differentiable function. Rather than leaving one word
out at a time,
we can simulate a word's removal by approximating
$f$ with a function that is linear in $w_i$ through the first-order Taylor expansion. 
The importance of $w_i$ is computed as the derivative of $f$ with respect to the one-hot vector:
\begin{equation}
    \frac{\partial f}{\partial w_i} \
    = \frac{\partial f}{\partial \bm{v}_i}\frac{\partial \bm{v}_i}{\partial w_i} \ 
    = \frac{\partial f}{\partial \bm{v}_i} \cdot \bm{v}_i 
\end{equation}

Thus, a word's importance is the dot product between the gradient of the class prediction
with respect to the embedding and the word embedding itself.
This gradient approximation simulates the change in
confidence when an input word is removed and has been used in various interpretation
methods for \abr{nlp}~\cite{arras2016explaining,
ebrahimi2017hotflip}. We refer to this interpretation approach as \textit{Gradient}
in our experiments. 

\subsection{Interpretation Method Failures}

Interpreting neural networks can have unexpected negative results. 
\citet{ghorbani2017interpretation} and \citet{kindermans2017unreliability}
show how a lack of model robustness and stability 
can cause egregious interpretation failures in computer vision settings.
\citet{feng2018rawr} extend this to \abr{nlp} and draw
connections between interpretation failures and adversarial examples~\cite{szegedy2013intriguing}.
To counteract this, new interpretation methods alone
are not enough---models must be improved. For instance, 
\citet{feng2018rawr} argue that interpretation methods should not rely
on prediction confidence as it does not reflect a model's uncertainty.

Following this, we improve interpretations by replacing
the softmax confidence with a more robust uncertainty estimate
using \dknn{}~\cite{papernot2018dknn}. This algorithm maintains
the accuracy of standard image classification models
while providing a better uncertainty metric capable of defending against adversarial examples.
\section{Deep k-Nearest Neighbors for Sequential Inputs}
\label{sec:deepknn}

This section describes Deep k-Nearest Neighbors,
its application to sequential inputs, and how we use it to 
determine word importance values.

\subsection{Deep k-Nearest Neighbors}

\citet{papernot2018dknn} propose Deep k-Nearest Neighbors
(\dknn{}), a modification to the test-time behavior
of neural networks.  

After training completes, the \dknn{} algorithm passes every training example
through the model and saves each of the layer's representations.
This creates a new dataset, whose features are the
representations and whose labels are the model predictions. Test-time predictions
are made by passing an example through the model and performing k-nearest neighbors classification
on the resulting representations. This modification does not degrade
the accuracy of image classifiers on several standard datasets~\cite{papernot2018dknn}.

For our purposes, the benefit of \dknn{} is the algorithm's uncertainty metric,
the \textit{conformity} score.  This score is the percentage of nearest
neighbors belonging to the predicted class. Conformity follows from the
framework of conformal prediction~\cite{shafer2008tutorial} and estimates how
much the training data supports a classification decision.  

The conformity score uses the representations at each neural network layer, and therefore, a prediction
only receives high conformity if it largely agrees with the training data at all representation
levels. This mechanism defends against adversarial examples~\cite{szegedy2013intriguing},
as it is difficult to construct a perturbation which changes the neighbors at every layer.
Consequently, conformity is a better uncertainty metric for both regular examples
and out-of-domain examples such as noisy or adversarial inputs,
making it suitable for interpreting models.

\subsection{Handling Sequences}

The \dknn{} algorithm requires fixed-size vector representations. 
To reach a fixed-size representation for text classification,
we take either the final hidden state of a
recurrent neural network or use max pooling across
time~\cite{collobert2008unified}. We consider deep architectures
of these two forms, using each of the layers' representations as the 
features for \dknn{}.

\subsection{Conformity \loo{}}
\label{sec:cred_l1o}

Using conformity, we generate interpretations through a modified
version of \loo{}~\cite{li2016understanding}. After removing a word,
rather than observing the drop in confidence, we instead
measure the drop in conformity. Formally, we
modify classifier $f$ in Equation \ref{eq:importance} to
output probabilities based on conformity. 
We refer to this method as \textit{conformity \loo{}}.

\section{\dknn{} Maintains Classification Accuracy}
\label{sec:experiments}

Interpretability should not come at the cost of performance---before investigating
how interpretable \dknn{} is, we first evaluate its accuracy. We
experiment with six text classification tasks and two models,
verifying that \dknn{} achieves accuracy comparable to regular classifiers.

\subsection{Datasets and Models}

We consider six common text classification tasks: binary sentiment analysis
using Stanford Sentiment Treebank~\cite[\abr{sst}]{socher2013recursive} and
Customer Reviews~\cite[\abr{cr}]{hu2004mining}, topic classification using
\emph{TREC} ~\cite{li2002trec}, opinion
polarity~\cite[\abr{mpqa}]{wiebe2005mpqa}, and
subjectivity/objectivity~\cite[\abr{subj}]{pang2004subj}. Additionally, we
consider natural language inference with \abr{snli}~\cite{bowman2015snli}.
We experiment with \bilstm{} and \cnn{} models.

\paragraph{\cnn{}}
Our \cnn{} architecture resembles \citet{kim2014convolutional}. We
use convolutional filters of size three, four, and five, with max-pooling over
time~\cite{collobert2008unified}. The filters are followed by
three fully-connected layers. We fine-tune \abr{GloVe}
embeddings~\cite{pennington2014glove} of each word. For \dknn{}, we use the
activations from the convolution layer and the three fully-connected layers.

\paragraph{\bilstm{}}
Our architecture uses a bidirectional \lstm{}~\cite{graves2005framewise}, with
the final hidden state forming the fixed-size
representation. We use three \lstm{} layers, followed by
two fully-connected layers. We fine-tune \abr{GloVe}
embeddings of each word. For \dknn{},
we use the final activations of the three recurrent layers and
the two fully-connected layers.

\paragraph{\abr{snli} Classifier}
Unlike the other tasks which consist of a single input sentence, \abr{snli} has two inputs,
a premise and hypothesis. Following \citet{conneau2017supervised}, we use the
same model to encode the two inputs, generating representations $u$ for the
premise and $v$ for the hypothesis. We concatenate these two representations along with
their dot-product and element-wise absolute difference, arriving at a final
representation $\left[u;v;u*v;\vert u-v\vert \right]$. 
This vector passes through two fully-connected
layers for classification. For \dknn{}, we use the
activations of the two fully-connected layers.

\paragraph{Nearest Neighbor Search}
For accurate interpretations, we trade efficiency for accuracy and replace locally
sensitive hashing~\cite{gionis1999lsh} used by
\citet{papernot2018dknn} with a k-d tree~\cite{bentley1975kdtree}. We use $k = 75$
nearest neighbors at each layer. The empirical results are robust to the choice
of $k$.

\subsection{Classification Results}

\dknn{} achieves comparable accuracy on the five classification tasks
(Table~\ref{table:prediction_accuracy}). On \abr{snli}, the \bilstm{} achieves
an accuracy of 81.2\% with a softmax classifier and 81.0\% with \dknn{}.

\begin{table*}[t]
\centering
\begin{tabular}{l|cccccc}
\toprule
& \abr{sst} & \abr{cr} & \abr{trec} & \abr{mpqa} & \abr{subj} \\
\midrule
\lstm{} & 86.7 & 82.7 & 91.5 & 88.9 & 94.8 \\
\lstm{} \dknn{} & 86.6 & 82.5 & 91.3 & 88.6 & 94.9 \\
\cnn{} & 85.7 & 83.3 & 92.8 & 89.1 & 93.5 \\
\cnn{} \dknn{} & 85.8 & 83.4 & 92.4 & 88.7 & 93.1 \\
\bottomrule
\end{tabular}
\caption{Replacing a neural network's softmax classifier with \dknn{} maintains classification accuracy on standard text classification tasks.}   
\label{table:prediction_accuracy}
\end{table*}
\section{\dknn{} is Interpretable}
\label{sec:interpretation}

Following past work~\cite{li2016understanding, murdoch2018cd}, we focus on the
\abr{sst} dataset for generating interpretations. Due to the lack of standard
interpretation evaluation metrics~\cite{doshivelez2017towards},
we use qualitative evaluations~\cite{smilkov2017smoothgrad,
sundararajan2017axiomatic, li2016understanding}, performing
quantitative experiments where possible to examine the distinction
between the interpretation methods.

\subsection{Interpretation Analysis}

\begin{table*}[t]
\centering
\begin{tabular}{lp{0.6\textwidth}}
\toprule
\textbf{Method} & \textbf{Saliency Map} \\
\midrule
Conformity &  \mybox{color5}{an} \mybox{color1}{intelligent} \mybox{color5}{fiction} \mybox{color5}{about} \mybox{color5}{learning} \mybox{color5}{through} \mybox{color5}{cultural} \mybox{color5}{clash}. \\
Confidence &  \mybox{color5}{an} \mybox{color1}{intelligent} \mybox{color6}{fiction} \mybox{color5}{about} \mybox{color5}{learning} \mybox{color5}{through} \mybox{color5}{cultural} \mybox{color9}{clash}. \\
Gradient &  \mybox{color5}{an} \mybox{color2}{intelligent} \mybox{color8}{fiction} \mybox{color5}{about} \mybox{color5}{learning} \mybox{color5}{through} \mybox{color5}{cultural} \mybox{color8}{clash}. \\
\midrule
Conformity &  \mybox{color5}{$<$Schweiger$>$} \mybox{color5}{is} \mybox{color2}{talented} \mybox{color5}{and} \mybox{color5}{terribly} \mybox{color2}{charismatic}. \\
Confidence &  \mybox{color6}{$<$Schweiger$>$} \mybox{color5}{is} \mybox{color2}{talented} \mybox{color5}{and} \mybox{color8}{terribly} \mybox{color2}{charismatic}. \\
Gradient &  \mybox{color6}{$<$Schweiger$>$} \mybox{color5}{is} \mybox{color2}{talented} \mybox{color5}{and} \mybox{color9}{terribly} \mybox{color2}{charismatic}. \\
\midrule
Conformity &  \mybox{color5}{Diane} \mybox{color5}{Lane} \mybox{color2}{shines} \mybox{color5}{in} \mybox{color5}{unfaithful}. \\
Confidence &  \mybox{color2}{Diane} \mybox{color5}{Lane} \mybox{color2}{shines} \mybox{color5}{in} \mybox{color6}{unfaithful}. \\
Gradient &  \mybox{color4}{Diane} \mybox{color5}{Lane} \mybox{color2}{shines} \mybox{color5}{in} \mybox{color8}{unfaithful}. \\
\bottomrule

\end{tabular}
\vspace{8pt}
\begin{tabular}{c}
Color Legend\quad  \mybox{color2}{Positive Impact}\quad \mybox{color7}{Negative Impact}
\end{tabular}
\caption{Comparison of interpretation approaches on \abr{sst} test
examples for the \bilstm{} model. Blue indicates positive impact and red
indicates negative impact. Our method (\textit{Conformity}
\loo{}) has higher precision, rarely assigning
importance to extraneous words such as ``clash'' or ``fiction''.}
\label{table:saliency_maps}
\end{table*}

We compare our method (\emph{Conformity \loo{}}) against two baselines: \loo{}
using regular confidence (\emph{Confidence \loo{}}, see Section \ref{sec:l10}) and the gradient with respect to the
input (\emph{Gradient}, see Section \ref{sec:gradient}). To create saliency maps, we normalize each word's importance
by dividing it by the total importance of the words in
the sentence. We display unknown words in angle brackets $<>$.
Table~\ref{table:saliency_maps} shows \abr{sst} interpretation examples
for the \bilstm{} model and further examples are shown on a
supplementary website.\footnote{\url{https://sites.google.com/view/language-dknn/}}

Conformity \loo{} assigns concentrated importance values to
a small number of input words. In contrast, the baseline methods
assign non-zero importance values to numerous words, many of which
are irrelevant.
For instance, in all three examples of Table~\ref{table:saliency_maps}, 
both baselines highlight almost half of the input, including words such as
``fiction'' and ``clash''. We suspect model confidence is oversensitive to
these unimportant input changes,
causing the baseline interpretations to highlight unimportant words. On the other hand, 
the conformity score better separates word importance, generating clearer interpretations.

The tendency for confidence-based approaches to assign importance 
to many words holds for the entire test set. We compute the average number of
highlighted words using a threshold of $0.05$ 
(a normalized importance value corresponding to a \mybox{color4}{light blue} or \mybox{color6}{light red} highlight). Out
of the average 20.23 words in \abr{sst} test set, gradient highlights 5.32 words,
confidence \loo{} highlights 5.79 words, and conformity \loo{} highlights 3.65 words.

The second, and related, observation for confidence-based approaches is a bias
towards selecting word importance based on a word's inherent sentiment,
rather than its meaning in context. For example, see ``clash'',
``terribly'', and ``unfaithful'' in Table \ref{table:saliency_maps}.
The removal of these words causes a small change in the model confidence.
When using \dknn{}, the conformity score indicates that the model's
uncertainty has not risen without these input words and \loo{} does not
assign them any importance.

We characterize our interpretation method as significantly
higher precision, but slightly lower recall than confidence-based
methods. Conformity \loo{} rarely assigns high importance to words
that do not align with
human perception of sentiment. However, there are cases when our
method does not assign significant importance to any word. This occurs
when the input has a high redundancy. For example, a positive movie
review that describes the sentiment in four distinct ways. In these
cases, leaving out a single sentiment word has little effect on the
conformity as the model's representation remains supported by the
other redundant features. Confidence-based interpretations, which
interpret models using the linear units that produce class scores,
achieve higher recall by responding to every change in the input for a
certain direction but may have lower precision.

In the second example of Table~\ref{table:saliency_maps}, the word ``terribly''
is assigned a negative importance value, disregarding its positive meaning in context.
To examine if this is a stand-alone example or a more general pattern of 
uninterpretable behavior, we calculate the importance value of the word
``terribly'' in other positive examples. For each occurrence of the word ``great''
in positive validation examples, we paraphrase it to ``awesome'', ``wonderful'',
or ``impressive'', and add the word ``terribly'' in front of it. 
This process yields $66$ examples. For each of these examples,
we compute the importance value of each input word and rank them from
most negative to most positive (the most negative word has
a rank of 1). We compare the average ranking of ``terribly''
from the three methods: 7.9 from conformity \loo{}, 1.68 from confidence \loo{}, and 1.1 from
gradient. The baseline methods consistently rank ``terribly'' as the most
negative word, ignoring its meaning in context. This echoes our suspicion: \dknn{} generates
interpretations with higher precision because conformity is robust to irrelevant input changes.
%(39.6\%)
%(8\%)
%(5\%)
\subsection{Analyzing Dataset Annotation Artifacts}
\label{sec:degenerate}

We use conformity \loo{} to interpret a model trained on \abr{snli}, a dataset known to
contain annotation artifacts. We demonstrate that our interpretation
method can help identify when models exploit dataset biases.

Recent studies~\cite{gururangan2018annotation,poliak2018hypothesis}
identify annotation artifacts in \abr{snli}. Superficial
patterns exist in the input which strongly correlate with certain
labels, making it possible for models to ``game'' the task: obtain high
accuracy without true understanding. For instance, the hypothesis of an
entailment example is often a general paraphrase of the premise, using
words such as ``outside'' instead of ``playing in a park''. Contradiction
examples often contain negation words or non-action verbs like ``sleeping''.
Models trained solely on the hypothesis can learn these patterns and reach
accuracies considerably higher than the majority baseline.

These studies indicate that the \abr{snli} task can be gamed. We look to confirm
that some artifacts
are indeed exploited by normally trained models that use full input pairs. We
create saliency maps for examples in the validation set using conformity
\loo{}. Table~\ref{table:annotation_artifacts} shows samples and more can be found on the
supplementary website. We use
blue highlights to indicate words which positively support the model's predicted class,
and the color red to indicate words that support a different class.
The first example is a randomly sampled baseline, showing how the words
``swims'' and ``pool'' support the model's prediction of contradiction.
The other examples are selected because they contain terms identified as artifacts.
In the second example, conformity \loo{} assigns extremely high word importance
to ``sleeping'', disregarding the other words necessary to predict contradiction (i.e., the neutral class is still possible
if ``pets'' is replaced with ``people''). In the final two hypotheses, the interpretation
method diagnoses the model failure, assigning high importance to ``wearing'', rather than
focusing positively on the shirt color.

To explore this further, we analyze the hypotheses in each \abr{snli} class
which contain a top five artifact identified by \citet{gururangan2018annotation}.
For each of these examples,
we compute the importance value for each input word using
both confidence and conformity \loo{}. We then rank the words from
most important for the prediction to least important (a score of 1
indicates highest importance) and report the average rank for the artifacts in Table~\ref{table:words}.
We sort the words by their Pointwise Mutual Information with the correct label as provided by \citet{gururangan2018annotation}.
The word ``nobody'' particularly stands out: it is the most important input word every time it appears
in a contradiction example. 

For most of the artifacts, conformity \loo{} assigns them a high importance,
often ranking the artifacts as the most important input word. 
Confidence \loo{} correlates less strongly with the known artifacts, frequently ranking them
as low as the fifth or sixth most important word. Given the high correlation between 
conformity \loo{} and the manually identified artifacts, this interpretation method may serve as
a technique to identify undesirable biases a model has learned.

\begin{table*}[t]
\centering
\begin{tabular}{llp{0.6\textwidth}}
\toprule
\textbf{Prediction} & \textbf{Input} & \textbf{Saliency Map} \\
\midrule
\multirow{2}{*}{Contradiction} & Premise & a young boy reaches for and touches the propeller of a vintage aircraft.  \\
& Hypothesis &  \mybox{color5}{a} \mybox{color8}{young} \mybox{color6}{boy} \mybox{color2}{swims} \mybox{color4}{in} \mybox{color6}{his} \mybox{color4}{pool}. \\
\midrule
\multirow{2}{*}{Contradiction} & Premise & a brown dog and a black dog in the edge of the ocean with a wave under them boats are on the water in the background.  \\
& Hypothesis &  \mybox{color5}{the} \mybox{color5}{pets} \mybox{color5}{are} \mybox{color2}{sleeping} \mybox{color5}{on} \mybox{color5}{the} \mybox{color5}{grass}. \\
\midrule
& Premise & man in a blue shirt standing in front of a structure painted with geometric designs.\\
Entailment & Hypothesis &  \mybox{color5}{a} \mybox{color6}{man} \mybox{color5}{is} \mybox{color2}{wearing} \mybox{color5}{a} \mybox{color6}{blue} \mybox{color5}{shirt}.\\
\textcolor{red}{Entailment} & Hypothesis &  \mybox{color5}{a} \mybox{color6}{man} \mybox{color5}{is} \mybox{color2}{wearing} \mybox{color5}{a} \mybox{color6}{black} \mybox{color5}{shirt}.\\
\bottomrule

\end{tabular}
\vspace{8pt}
\begin{tabular}{c}
Color Legend\quad  \mybox{color2}{Positive Impact}\quad \mybox{color7}{Negative Impact}
\end{tabular}
\caption{Interpretations generated with conformity \loo{} align with
annotation biases identified in \abr{snli}. In the second example, the
model puts emphasis on the word ``sleeping'', disregarding other words
that could indicate the Neutral class. The final example diagnoses a
model's incorrect Entailment prediction (shown in red). Green
highlights indicate words that support the classification decision
made (shown in parenthesis), pink highlights indicate words that
support a different class.}
\label{table:annotation_artifacts}
\end{table*}

\begin{table}
\setlength{\tabcolsep}{2pt}
\centering
\small
\begin{tabular}{clcc}
\toprule
\textbf{Label} & \textbf{Artifact} & \textbf{Conformity} & \textbf{Confidence} \\
\midrule
\multirow{5}{*}{\textbf{Entailment}}
& outdoors & 2.93 & 3.26 \\
& least & 2.22 & 4.41 \\
& instrument & 3.57 & 4.47 \\
& outside & 4.08 & 4.80 \\
& animal & 2.00 & 4.73 \\
\midrule
\multirow{5}{*}{\textbf{Neutral}}
& tall & 1.09 & 2.61 \\
& first & 2.14 & 2.99 \\
& competition & 2.33 & 5.56 \\
& sad & 1.39 & 1.79 \\
& favorite & 1.69 & 3.89 \\
\midrule
\multirow{5}{*}{\textbf{Contradiction}}
& nobody & 1.00 & 1.00 \\
& sleeping & 1.64 & 2.34 \\
& no & 2.53 & 5.74 \\
& tv & 1.92 & 3.74 \\
& cat & 1.42 & 3.62 \\
\bottomrule
\end{tabular}
\caption{The top \abr{snli} artifacts identified by \citet{gururangan2018annotation} are shown on the left. For each word, we compute the average importance rank over the validation set using either \textit{Conformity} or \textit{Confidence} \loo{}. A score of 1 indicates that a word is always ranked as the most important word in the input. \textit{Conformity} \loo{} assigns higher importance to artifacts, suggesting it better diagnoses model biases.}
\label{table:words}
\end{table}

\section{Discussion and Related Work}
\label{sec:discussion}

We connect the improvements made by conformity \loo{} to 
model confidence issues, compare alternative interpretation improvements,
and discuss further features of \dknn{}.

\subsection{Issues in Neural Network Confidence}
\label{sec:overconfidence}

Many existing feature attribution methods rely on estimates
of model uncertainty: both input gradient and confidence \loo{} rely on
prediction confidence, our method relies on \dknn{} conformity.
Interpretation quality is thus determined by reliable uncertainty estimation. 
For instance, past work shows relying on neural network confidence
can lead to unreasonable interpretations~\cite{kindermans2017unreliability,
ghorbani2017interpretation, feng2018rawr}. Independent of interpretability,
\citet{guo2017calibration} show that neural network confidence is unreasonably
high: on held-out examples, it far exceeds empirical accuracy.
This is further exemplified by the high confidence predictions produced on
inputs that are adversarial~\cite{szegedy2013intriguing} or contain solely
noise~\cite{goodfellow2014explaining}.

\subsection{Confidence Calibration is Insufficient}

We attribute one interpretation failure to neural network
confidence issues. \citet{guo2017calibration} study
overconfidence and propose a calibration procedure using
Platt scaling, which adjusts the temperature parameter of the softmax
function to align confidence with accuracy on a held-out dataset.
However, this is not input dependent---the confidence is lower
for both full-length examples and ones with words left out. 
Hence, selecting influential words will remain difficult.

To verify this, we create an interpretation baseline using temperature 
scaling.
The results corroborate the intuition: calibrating the confidence 
of \loo{} does not improve interpretations. Qualitatively, the 
calibrated interpretation results remain comparable to  
confidence \loo{}. Furthermore, calibrating the \dknn{}
 conformity score as in \citet{papernot2018dknn} did not improve
interpretability compared to the uncalibrated conformity score.

\subsection{Alternative Interpretation Improvements}

Recent work improves interpretation methods through other
means. \citet{smilkov2017smoothgrad} and \citet{sundararajan2017axiomatic}
both aggregate gradient values over multiple
backpropagation passes to eliminate local noise or satisfy
interpretation axioms. This work does not address 
model confidence and is orthogonal to our \dknn{} approach. 
 
\subsection{Interpretation Through Data Selection}

Retrieval-Augmented Convolutional Neural Networks~\cite{zhao2018retrieval} are similar
to \dknn{}: they augment model predictions with an information
retrieval system that searches over network activations from the training data. 

Retrieval-Augmented models and \dknn{} can both select influential
training examples for a test prediction. In particular, the
training data activations which are closest to the test point's 
activations are influential according to the
model. These training examples
can provide interpretations as a form of analogy~\cite{caruana1999casebased}, an
intuitive explanation for both machine learning experts and non-experts~\cite{klein1989biases,
kim2014bayesian,
koh2017influence, wallace2018trick}. However, unlike in computer vision where
training data selection using \dknn{} yielded interpretable examples~\cite{papernot2018dknn},
our experiments did not find human interpretable data points for \abr{sst} or
\abr{snli}. 
  		
\subsection{Trust in Model Predictions}

Model confidence is important for real-world
applications: it signals how much one should trust
a neural network's predictions. Unfortunately, users may be 
misled when a model outputs highly confident predictions on rubbish
examples~\cite{goodfellow2014explaining, nguyen2015fooled} or adversarial
examples~\cite{szegedy2013intriguing}. Recent work decides when
to trust a neural network model~\cite{ribeiro2016lime,
doshivelez2017towards, jiang2018trust}. For instance, analyzing local linear
model approximations~\cite{ribeiro2016lime} or flagging rare network activations
using kernel density estimation~\cite{jiang2018trust}. The \dknn{} conformity
score is a trust metric that helps defend against image adversarial
examples~\cite{papernot2018dknn}. Future work should study if this robustness
extends to interpretations. 

\section{Future Work and Conclusion}
\label{sec:conclusion}

A robust estimate of model uncertainty is critical to determine
feature importance. The \dknn{} conformity score is
one such uncertainty metric which leads to higher precision interpretations.
Although \dknn{} is only a test-time improvement---the model is
still trained using maximum likelihood. Combining nearest
neighbor and maximum likelihood objectives during training may further
improve model accuracy and
interpretability. Moreover, other uncertainty estimators
do not require test-time modifications. For
example, modeling $p(x)$ and $p(y\mid\mb{x})$ using Bayesian Neural
Networks~\cite{gal2016uncertainty}.

Similar to other \abr{nlp} interpretation
methods~\cite{sundararajan2017axiomatic, li2016understanding}, conformity
\loo{} works when a model's representation has a fixed size. For
other \abr{nlp} tasks, such as structured prediction (e.g., translation and
parsing) or span prediction (e.g., extractive summarization and reading
comprehension), models output a variable number of predictions and our interpretation
approach will not suffice. Developing interpretation techniques for these
types of models is a necessary area for future work.

We apply \dknn{} to neural models for text classification. This provides a
better estimate of model uncertainty---conformity---which we combine with
\loo{}. This overcomes issues stemming from neural network confidence,
leading to higher precision interpretations. Most interestingly,
our interpretations are supported by the training data,
providing insights into the representations learned by a model.

\section*{Acknowledgments}
Feng was supported under subcontract to Raytheon BBN Technologies by DARPA award
HR0011-15-C-0113.
JBG is supported by NSF Grant IIS1652666.
Any opinions, findings, conclusions, or recommendations expressed here are those
of the authors and do not necessarily reflect the view of the sponsor.
The authors would like to thank the members of the CLIP lab at the University of Maryland
and the anonymous
reviewers for their feedback.

\bibliography{journal-abbrv,fs,jbg}
\bibliographystyle{emnlp_natbib}

\end{document}